\documentclass[pamm,a4paper,fleqn]{w-art}
\usepackage[nameinlink]{cleveref}
\usepackage{times,cite,w-thm}
\usepackage[T1]{fontenc}
\usepackage[utf8]{inputenc}
\theoremstyle{plain}

\theoremstyle{definition}

\usepackage{graphicx}
\usepackage[american]{babel}

\usepackage{graphicx}

\usepackage{amssymb}
\usepackage{amsmath}

\usepackage{multirow}

\usepackage[vlined,longend,linesnumbered,ruled]{algorithm2e}
\usepackage{mathabx}

\usepackage{subcaption}

\usepackage{comment}

\newcommand{\R}{\mathbb{R}}
\newcommand{\N}{\mathbb{N}}
\newcommand{\norm}[1]{\left\lVert#1\right\rVert}
\newcommand{\normC}[1]{\left\lVert#1\right\rVert_{C(W;\mathbb{R}^n)}}
\newcommand{\normCr}[1]{\left\lVert#1\right\rVert_{C^r(W;\mathbb{R}^n)}}
\newcommand{\normt}[1]{\left\lVert#1\right\rVert_{C^r([0,T]\times W;\mathbb{R}^n)}}
\newcommand{\normto}[1]{\left\lVert#1\right\rVert_{C^{0,r}([0,T]\times W;\mathbb{R}^n)}}
\newcommand{\normtod}[1]{\left\lVert#1\right\rVert_{C^{0,r}([0,T]\times W_\Phi;\mathbb{R}^{2d})}}
\newcommand{\normtd}[1]{\left\lVert#1\right\rVert_{C^r([0,T]\times W;\mathbb{R}^{2d})}}
\newcommand{\normCrmax}[1]{\left\lVert#1\right\rVert_{C_{\max}^r(W;\mathbb{R}^n)}}
\newcommand{\normtmax}[1]{\left\lVert#1\right\rVert_{C_{\max}^r([0,T]\times W;\mathbb{R}^n)}}
\newcommand{\strich}{\,\middle\vert\,}
\newcommand{\abs}[1]{\left\lvert#1\right\rvert}
\newcommand{\pq}{\begin{pmatrix}p \\ q \end{pmatrix}}

\DeclareMathOperator*{\id}{id}
\DeclareMathOperator*{\diag}{diag}
\DeclareMathOperator*{\dist}{dist}

\usepackage{aliascnt}
\newaliascnt{definition}{theorem}
\newtheorem{definition}[definition]{Definition}
\aliascntresetthe{definition}

\crefname{definition}{Definition}{Definitions}
\Crefname{definition}{Definition}{Definitions}

\begin{document}
%\renewenvironment{definition}{\definitionname}[section]
%% \def\leftmark{Session title}
%%
%%    The information for the title page will be placed between
%%    \begin{document} and \maketitle. The order of most entries
%%    is determined by the class file and can not be changed by
%%    rearranging them. The maketitle command follows after the
%%    abstract.
%%
%%    Most of the following commands will be completed by the publisher.
%%
    \renewcommand{\copyrightyear}{2025}
%%    \DOIsuffix{pamm.20161zzzz}
%%    \Volume{16} 
    \Year{2025} 
%%    \pagespan{1}{}
%%
%%    The short title is optional:

\TitleLanguage[EN]
\title[TH\'enonNets]{Time-adaptive H\'enonNets for separable Hamiltonian systems}

%% Please do not enter footnotes or \inst{}-notes into the optional
%% argument of the author command. 

%% Please delete not needed author entries.
%% Information for the first author.
\author{\firstname{Konrad}  \lastname{Janik}\inst{1,}%
  \footnote{Corresponding author: e-mail \ElectronicMail{janik@mpi-magdeburg.mpg.de}}}

\address[\inst{1}]{\CountryCode[DE]Max Planck Institute for Dynamics of Complex Technical Systems, Sandtorstra\ss{}e 1, 39106 Magdeburg}
%%
%%    Information for the second author
\author{\firstname{Peter} \lastname{Benner}\inst{1,2,}%
  }

\address[\inst{2}]{\CountryCode[DE]Otto von Guericke University Magdeburg,  Fakultät für Mathematik, Universitätsplatz 2, 39106 Magdeburg, Germany}
%%

%%
%%    \dedicatory{This is a dedicatory.}
%%
%%    Abstract is required.
\AbstractLanguage[EN]
\begin{abstract}
  Measurement data is often sampled irregularly, i.e., not on equidistant time grids. This is also true for Hamiltonian systems. However, existing machine learning methods, which learn symplectic integrators, such as SympNets \cite{JinZZetal20} and HénonNets \cite{BurTM21} still require training data generated by fixed step sizes. To learn time-adaptive symplectic integrators, an extension to SympNets called TSympNets is introduced in \cite{JanB25}. The aim of this work is to do a similar extension for HénonNets.  We propose a novel neural network architecture called T-HénonNets, which is symplectic by design and can handle adaptive time steps. We also extend the T-HénonNet architecture to non-autonomous Hamiltonian systems. Additionally, we provide universal approximation theorems for both new architectures for separable Hamiltonian systems and discuss why it is difficult to handle non-separable Hamiltonian systems with the proposed methods. To investigate these theoretical approximation capabilities, we perform different numerical experiments.
\end{abstract}
%% maketitle must follow the abstract.
\maketitle                   % Produces the title.

\section{Introduction}
The numerical integration of Hamiltonian systems is a central topic in computational physics and theoretical chemistry. Concrete applications are found in celestial mechanics, molecular design, and many more \cite{LeiR05,HaiLW06,Arn13}. Though the topic has been well-researched, particularly since the 1990s, when many structured integration schemes were developed, novel interest arose with the advent of machine learning and artificial intelligence for scientific computing. This allows, in particular, deriving compact Hamiltonian models from data — be it experimental or computed via simulation using a legacy code — by learning the Hamiltonian \cite{GreDY19,BerDMetal19} or the Lagrangian in the position-velocity space \cite{CranGHetal20}. There are many extensions to these Hamiltonian neural networks (HNNs) \cite{CheZAetal20, DavM23, HorSKetal25, XioTHetal21, YilGBetal23, HanGH21}. While the original approach still required derivative data, the neural ODE method \cite{CheRBetal18} was quickly implemented with symplectic integrators to preserve the structure of the Hamiltonian system \cite{DavM23, CheZAetal20}. Because of the implicit nature of symplectic integration schemes \cite{HaiLW06}, the fast evaluation of the learned models was limited to separable Hamiltonian systems (H(p,q)=K(p)+V(q)). To tackle this problem, phase space extensions were introduced in \cite{Tao16, XioTHetal21}. Additionally, an approach on how to handle parametric Hamiltonian systems was given in \cite{HanGH21}. Another way of learning Hamiltonian systems is the identification of proper generating functions with neural networks \cite{CheT21}.
Yet another approach consists of learning a suitable symplectic integrator for Hamiltonian systems using deep learning, namely SympNets \cite{JinZZetal20}, HénonNets \cite{BurTM21}, and generalized Hamiltonian neural networks \cite{HorSKetal25}. SympNets are designed as a composition of different shear maps, while HénonNets are compositions of Hénon-like maps \eqref{eqn:henonmap}. Both of these architectures fall into the framework of generalized HNNs as compositions of symplectic integrators evaluated for different separable Hamiltonians in each layer. All of these methods are universal approximators of symplectic maps. For SympNets, a time-adaptive extension was proposed in \cite{JanB25}. This work aims to propose a time-adaptive HénonNet architecture and discuss their approximation capabilities. Additionally, we extend these time-adaptive HénonNets to non-autonomous Hamiltonian systems, where the Hamiltonian (H(p,q,t)) explicitly depends on time.\\
The rest of the paper is structured as follows. First, we introduce some necessary notation in \Cref{subsec:notation} and briefly recap Hamiltonian systems in \Cref{subsec:ham}. In \Cref{sec:henon}, we introduce our new neural network architectures before we discuss their theoretical properties in \Cref{sec:theory}, specifically which Hamiltonian systems they can (\Cref{subsec:uats}) and cannot (\Cref{subsec:lim}) be used for. In \Cref{sec:exp}, we show that our theoretical findings also have practical implications by testing the previously discussed methods on three different examples, i.e., the mathematical pendulum in \Cref{subsec:pend}, a linear but non-separable example in \Cref{subsec:lin}, and the harmonic oscillator with an external force as an example of a non-autonomous Hamiltonian system in \Cref{subsec:ho}. Lastly, we sum up our results in \Cref{sec:conc} and discuss future research possibilities.
\subsection{Notation}
\label{subsec:notation}
In this subsection, we introduce some notation that will be useful later.
Let $d ,n \in \N, r \in \N_0, W \subseteq \R^d$ be a compact set. We define the norm on $C^r(W;\R^n)$ by
\begin{align*}
	\normCr{f}:= \sum_{\abs{\alpha}\leq r} \max_{1 \leq i \leq n} \sup_{x \in W}\abs{D^\alpha f^{(i)}(x)},
\end{align*}
with
\begin{align*}
	D^\alpha f := \frac{\partial^{\abs{\alpha}}f}{\partial x_1^{\alpha_1} \dotsm \partial x_d^{\alpha_d}}
\end{align*}
and $f=(f^{(1)},...,f^{(n)})^T.$ As is usually done, we just write $C^r(W)$ for $n=1.$ 
When we are looking at time-dependent functions $f:[0,T]\times W \to \R^n$ with $T>0,$ we will be using the standard $C^r$ norm given by
\begin{align*}
    \normt{f}:=\sum_{\abs{\alpha}\leq r} \max_{1\leq i \leq n} \sup_{(t,x) \in [0,T]\times W} \abs{D^\alpha f^{(i)}(t,x)},
\end{align*}
where $D^\alpha f$ might also contain time derivatives.
Furthermore, we want to denote closed balls around a set $M\subseteq \R^n$ with radius $R$ by
\begin{align*}
	\overline{{B}_R(M)}:=\left\{x \in \R^n \vert \exists y \in M: \norm{x-y}_\infty \leq R\right\}.
\end{align*}
The next notation we want to introduce is that of matrix-like functions, which we will use to define HénonNets.
As in \cite{JinZZetal20}, we will denote them with
\begin{align*}
	\begin{bmatrix}
	f_1 & f_2 \\
	f_3 & f_4
	\end{bmatrix}: \R^{2d} \to \R^{2d}, \qquad
	\begin{bmatrix}
	f_1 & f_2 \\
	f_3 & f_4
	\end{bmatrix}
	\begin{pmatrix}
	p\\q
	\end{pmatrix}:=
	\begin{pmatrix}
	f_1(p)+f_2(q)\\
	f_3(p)+f_4(q)
	\end{pmatrix},
\end{align*}
where $f_i:\R^{d} \to \R^d,~i=1,...,4$.\\
Lastly, we write the set of neural networks with one hidden layer and activation function $\sigma$ as
\begin{align*}
    \Sigma_d(\sigma)=\left\{f: \R^n \mapsto \R \strich f(x)=a^T \sigma(K x +b), a,b \in \R^m, K \in \R^{n\times m}, m \in \N\right\}
\end{align*}
like in \cite{HorSW90}.
\subsection{Hamiltonian systems}
\label{subsec:ham}
The problems we want to consider in this work are from the class of Hamiltonian systems.
Hamiltonian systems are given by
\begin{align}
	\label{eqn:Hsys}
	\dot{x}(t)= J^{-1} \nabla H(x(t), t), \qquad x(t_0)=x_0
\end{align}
with $x= (p^T, q^T)^T \in \R^{2d}$ and 
\begin{align*}
	J=\begin{pmatrix}
	0 & I_d\\
	-I_d & 0
	\end{pmatrix}\in \R^{2d \times 2d}.
\end{align*}
A Hamiltonian system \eqref{eqn:Hsys} is called autonomous if the Hamiltonian does not explicitly depend on time , i.e., $\partial_t H(x,t)=0$. Otherwise, it is called non-autonomous.\\
It is well known that the flow of a Hamiltonian system is symplectic \cite{HaiLW06}. Therefore, many modern machine learning methods that try to learn the flow of Hamiltonian systems aim to enforce symplecticity in one way or another \cite{JinZZetal20, BurTM21, JanB25, CheZAetal20, DavM23, CheT21, HorSKetal25}. HénonNets \cite{BurTM21} and SympNets \cite{JinZZetal20} provide a neural Network architecture that is symplectic by design. We aim to extend the HénonNet architecture to be time-adaptive and suitable for non-autonomous Hamiltonian systems, i.e., systems, where the Hamiltonian $H(x, t)$ explicitly depends on time, like SympNets \cite{JinZZetal20} were extended in \cite{JanB25}.
\section{HénonNets}
\label{sec:henon}
HénonNets are an intrinsically symplectic neural network architecture, which were introduced in \cite{BurTM21}. In this section, we first introduce the original HénonNets from \cite{BurTM21}. Then we extend them to be time-adaptive, and lastly, we propose a new architecture for non-autonomous Hamiltonian systems.
\subsection{Original HénonNets}
HénonNets are compositions of mappings similar to Hénon-like maps. The layers are given by a composition of four mappings like
\begin{align*}
    \mathcal{H}_{V_i,\eta_i}\begin{pmatrix}
        p \\ q
    \end{pmatrix}=
    \begin{pmatrix}
        \nabla V_i(p)-q\\
        p + \eta_i
    \end{pmatrix},
\end{align*}
with trainable neural networks $V_i$ and trainable parameters $\eta_i$. For the purpose of this work, we use fully-connected neural networks with one hidden layer, $V_i \in \Sigma_d(\sigma)$ and activation function $\sigma$, but other architectures are possible as well. This way we can analytically calculate the gradient, which saves time in the forward pass.
The potential $V_i$ and shift $\eta_i$ are the same throughout one layer, but can be different from layer to layer. 
\begin{definition}[HénonNets \cite{BurTM21}]
    \label{def:henonnet}
    Let
    \begin{align}
        \label{eqn:henonmap}
        \mathcal{H}_{V_i,\eta_i}\left(\begin{pmatrix}
            p \\ q
        \end{pmatrix}\right)=
        \begin{pmatrix}
            \nabla V_i(p)-q\\
            p + \eta_i
        \end{pmatrix},
    \end{align}
    with a fully-connected neural network with one hidden layer $V_i \in \Sigma_{d}(\sigma)$ and activation function $\sigma,$ and a trainable parameter $\eta_i\in \R^d$. A \textbf{Hénonlayer} is defined by applying the same Hénon-like map \eqref{eqn:henonmap} four times
    \begin{align*}
        \mathbf{H}_{V_i, \eta_i}=\mathcal{H}_{V_i, \eta_i}\circ\mathcal{H}_{V_i, \eta_i}\circ\mathcal{H}_{V_i, \eta_i}\circ\mathcal{H}_{V_i, \eta_i}.
    \end{align*}
    We define the set of \textbf{HénonNets} as finite compositions of Hénonlayers
    \begin{align*}
        \Psi_{\text{H}}:=\left\{\psi=\mathbf{H}_{V_{m},\eta_{m}} \circ \cdots \circ \mathbf{H}_{V_1,\eta_1} \strich m \in \N\right\}.
    \end{align*}
    %For actually building these maps into a neural network one can use any network architecture which is able to approximate $C^{r+1}$ functions.
\end{definition}
We can easily check that the maps defined in \eqref{eqn:henonmap} are symplectic. Using the fact that symplectic mappings form a group with composition as their group operation, we observe that all HénonNets are symplectic by design.\\
Furthermore, HénonNets can approximate any symplectic map arbitrarily well, since every symplectic map can be approximated by a composition of $4m,~m \in \N$, maps defined in \eqref{eqn:henonmap} \cite[Theorem 1]{Tur03}.

\subsection{Time-adaptive HénonNets}
\label{subsec:thenonnets}
Following the time-adaptive extension of SympNets \cite{JanB25}, the analogous idea would be to replace \eqref{eqn:henonmap} by
\begin{align}
    \label{eqn:naiv_henon}
    \mathcal{H}_{V,\eta}\left(h, \pq\right)=\begin{bmatrix}
        h\nabla V & -I\\ I & 0
    \end{bmatrix}\pq + \begin{pmatrix}
        0 \\ \eta
    \end{pmatrix}.
\end{align}
Unfortunately, this comes with some problems. We will show in \Cref{subsec:lim} that these T-HénonNets can only approximate Hamiltonian systems where the Hamiltonian is given by $H(p,q)= V(-p) + V(q)$. Even though it is not obvious, the solution to this problem is a rather simple change in the architecture given by the following definition.
\begin{definition}[T-HénonNets]
    \label{def:thenonnet}
    Let
    \begin{align}
        \label{eqn:thenonmap}
        \mathcal{H}_{V_i,\eta_i}\left(h,\begin{pmatrix}
            p \\ q
        \end{pmatrix}\right)=
        \begin{bmatrix}
            h \nabla V_i & -I \\ I & 0
        \end{bmatrix}\pq + \begin{pmatrix}
            \eta_i^{(p)}\\ \eta_i^{(q)}
        \end{pmatrix}
    \end{align}
    with a fully-connected neural network with one hidden layer $V_i \in \Sigma_{d}(\sigma)$ and activation function $\sigma,$ and a trainable parameter $\eta_i=((\eta_i^{(p)})^T, (\eta_i^{(q)})^T)^T\in \R^{2d}$. A \textbf{T-Hénonlayer} is defined by applying the same Hénon-like map \eqref{eqn:thenonmap} four times
    \begin{align*}
        \mathbf{H}_{V_i, \eta_i}(h,x)=(\mathcal{H}_{V_i, \eta_i}(h,\cdot)\circ\mathcal{H}_{V_i, \eta_i}(h,\cdot)\circ\mathcal{H}_{V_i, \eta_i}(h,\cdot)\circ\mathcal{H}_{V_i, \eta_i}(h,\cdot))(x).
    \end{align*}
    We define the set of \textbf{time-adaptive HénonNets (T-HénonNets)} as finite compositions of T-Hénonlayers
    \begin{align*}
        \Psi_{\text{TH}}:=\left\{\psi(h,x)=(\mathbf{H}_{V_{m},\eta_{m}}(h,\cdot) \circ \cdots \circ \mathbf{H}_{V_1,\eta_1}(h,\cdot))(x) \strich m \in \N\right\}.
    \end{align*}
\end{definition}
Note that the only difference to the T-HénonNets based on \eqref{eqn:naiv_henon} is that we add an additional bias $\eta^{(p)}$. Obviously, this change does not impact the symplecticity of the T-HénonNets. We will show in \Cref{subsec:uats} that the T-HénonNets from \Cref{def:thenonnet} are universal approximators for symplectic maps.

\subsection{Non-autonomous time-adaptive HénonNets}
Now we want to focus on non-autonomous Hamiltonian systems, i.e., $H(p,q,t)$. As discussed in \cite{JanB25}, it is possible to transform a non-autonomous Hamiltonian system into one that is autonomous. However, this phase space extension generally does not preserve the separability of the Hamiltonian system. Current explicit time-adaptive symplectic neural network architectures like TSympNets \cite{JanB25} and our T-HénonNets are not able to deal with non-separable Hamiltonian systems. Hence, we have to deal with the non-autonomous case instead of transforming the system into an autonomous, but non-separable Hamiltonian system.\\
Just like the non-autonomous extension of TSympNets in \cite{JanB25}, the idea for non-autonomous T-HénonNets is to allow the potentials $V_i$ to explicitly depend on time.
\begin{definition}[Non-autonomous T-HénonNets]
    \label{def:nathenonnet}
    Let
    \begin{align*}
        \mathcal{H}_{V_i,\eta_i}\left(h,t, \begin{pmatrix}
            p \\ q
        \end{pmatrix}\right)=
        \begin{bmatrix}
            h \nabla_x V_i(t, \cdot) & -I \\ I & 0
        \end{bmatrix}\pq + \begin{pmatrix}
            \eta_i^{(p)}\\ \eta_i^{(q)}
        \end{pmatrix}
    \end{align*}
    with a fully-connected neural network with one hidden layer $V_i \in \Sigma_{d+1}(\sigma)$ and activation function $\sigma,$ and a trainable parameter $\eta_i=((\eta_i^{(p)})^T, (\eta_i^{(q)})^T)^T\in \R^{2d}$. Additionally, let
    \begin{align}
        \label{eqn:nathenonmap}
        \mathcal{H}_{V_i,\eta_i}^{(m)}\left(h,t, \begin{pmatrix}
            p \\ q
        \end{pmatrix}\right)=\begin{pmatrix}
            t + \frac{h}{4m}\\
            \mathcal{H}_{V_i,\eta_i}\left(h,t, \pq\right)
        \end{pmatrix}
    \end{align}
    A \textbf{NAT-Hénonlayer} is defined by applying the same Hénon-like map \eqref{eqn:nathenonmap} four times
    \begin{align*}
        \mathbf{H}_{V_i, \eta_i}(h, t, x)=(\mathcal{H}_{V_i, \eta_i}^{(m)}(h, \cdot)\circ\mathcal{H}_{V_i, \eta_i}^{(m)}(h, \cdot)\circ\mathcal{H}_{V_i, \eta_i}^{(m)}(h, \cdot)\circ\mathcal{H}_{V_i, \eta_i}^{(m)}(h, \cdot))(t, x).
    \end{align*}
    We define the set of \textbf{non-autonomous, time-adaptive HénonNets (NAT-HénonNets)} as finite compositions of NAT-Hénon\-layers
    \begin{align*}
        \Psi_{\text{NATH}}:=\left\{\psi(h,t, x)=(\mathbf{H}_{V_{m},\eta_{m}}(h, \cdot) \circ \cdots \circ \mathbf{H}_{V_1,\eta_1}(h, \cdot))(t, x) \strich m \in \N\right\}.
    \end{align*}
\end{definition}
\section{Theory of T-HénonNets}
\label{sec:theory}
In this section, we will discuss some theory for the T-HénonNets defined in the previous section. We start with their approximation properties.
\subsection{Approximation capabilities}
\label{subsec:uats}
T-HénonNets are universal approximators for flows of separable Hamiltonian systems; see \Cref{the:henonnet}. To prove this result, we need some theorems from \cite{JanB25} that give conditions for composition-based approximations of ODE flows, namely \Cref{the:comp,the:compcor}. To make the upcoming proofs easier to follow, we state them here.
\begin{theorem}[\cite{JanB25}]
    \label{the:comp}
    Let $n, m, r\in \N,~ U \subseteq \R^n$ open, $f \in C^r(U;\R^n)$ and $\Phi:\R^+_0 \times U \to \R^n$ be the phase flow of the ODE
	\begin{align*}
		\dot{x}(t)=f(x), \qquad x(0)=x_0,
	\end{align*}
	i.e., $\Phi(t,x_0)=x(t)$. Let $T>0$ such that $\Phi(t,x)$ exists for all $t \in [0,T],~x \in U$ and $\Phi([0,T]\times U) \subseteq U$. We define $\Phi_{i,m}(h,x):=\Phi(ih/m,x)$. Furthermore, let $g_i \in C^r([0,T]\times\R^n;\R^n)$, $g_{i,m}(h,x):=g_i(h/m,x)$ and $G_{i,m}(h,x):=g_{i,m}(h,g_{i-1,m}(h,...g_{1,m}(h,x)...))$ for $i=1,...,m$. For a compact set $W \subseteq U$ we define $W_\Phi:=\Phi([0,T],W)$ and $\tilde{W}:=\overline{B_R(W_\Phi)}$. with $R < \dist(W_\Phi,\partial U)$.
	Let there be constant $C<\infty$ independent of $m$ such that
	\begin{align}
		\label{eqn:compcond}
		\norm{g_{i,m}-\Phi_{1,m}}_{C^r([0,T]\times \tilde{W};\R^n)}\leq \frac{C}{m^2}
	\end{align}
	for $i=1,...,m$. Then there is $m_0 \in \N$ such that for all $m \geq m_0$ the difference between the composition $G_{m,m}$ and the phase flow $\Phi$ decreases with $1/m$, i.e.,
	\begin{align}
		\label{eq:comp}
		\normt{\Phi-G_{m,m}}\leq \frac{\tilde{C}}{m},
	\end{align}
	where $\tilde{C}$ does not depend on $m$.
\end{theorem}
\begin{theorem}[\cite{JanB25}]
    \label{the:compcor}
	Let $n, m, r\in \N,~ U \subseteq \R^n$ open, $W_t \subseteq \R$ compact, $f \in C^r(U \times W_t;\R^n)$ and $\Phi:\R^+_0 \times W_t\times U \to \R^n$ be the phase flow of the ODE
	\begin{align}
		\label{eq:nauto}
		\dot{x}(t)=f(x,t), \qquad x(t_0)=x_0,
	\end{align}
	i.e., $\Phi(t,t_0,x_0)=x(t)$. Let $T>0$ such that $\Phi(t,t_0,x)$ exists for all $t \in [0,T],~t_0\in W_t,~x \in U$ and $\Phi([0,T]\times W_t\times U) \subseteq U$. We define $\Phi_{1,i,m}(h,t,x):=\Phi(h/m,t+ih/m,x)$. Furthermore, let $g_i \in C^r([0,T]\times \R\times\R^n;\R^n)$, $g_{i,m}(h,t,x):=g_i(h/m,t,x)$ and 
	\begin{align*}
		G_{i,m}(h,t,x):=g_{i,m}\left(h,t+\frac{(i-1)h}{m},g_{i-1,m}\left(h,t+\frac{(i-2)h}{m},...,g_{1,m}(h,t,x)...\right)\right)
	\end{align*}
	for $i=1,...,m$. For a compact set $W \subseteq U$ we define $W_\Phi:=\Phi([0,T],W_t,W)$ and $\tilde{W}:=\overline{B_R(W_\Phi)}$. with $R < \dist(W_\Phi,\partial U)$. Also, we define $\tilde{W}_t:=\left\{t_0+t \strich t_0 \in W_t,~t \in [0,T]\right\}$.
	Let there be constant $C<\infty$ independent of $m$ such that
	\begin{align*}
		\norm{g_{i,m}-\Phi_{1,i,m}}_{C^r([0,T]\times \tilde{W}_t \times \tilde{W};\R^n)}\leq \frac{C}{m^2}
	\end{align*}
	for $i=1,...,m$. Then there is $m_0 \in \N$ such that for all $m \geq m_0$ the difference between the composition $G_{m,m}$ and the phase flow $\Phi$ decreases with $1/m$, i.e.,
	\begin{align*}
		\norm{\Phi-G_{m,m}}_{C^r([0,T]\times W_t \times W;\R^{n})}\leq \frac{\tilde{C}}{m},
	\end{align*}
	where $\tilde{C}$ does not depend on $m$.
\end{theorem}
Finally, we need the following definition for activation functions to state our universal approximation theorems.
\begin{definition}[$r$-finite \cite{JinZZetal20}]
	Let $r \in \N_0.$ A function $\sigma \in C^r(\R)$ is called \textbf{$r$-finite} if
	\begin{align*}
		0< \int_\R \abs{D^r(\sigma)(x)} dx < \infty.
	\end{align*}
\end{definition}

\begin{theorem}[Approximation Theorem for T-HénonNets]
    \label{the:henonnet}
    For any $r \in \mathbb{N}$ and open $U \subseteq \R^{2d}$, the set of time-dependent HénonNets $\Psi_{\text{TH}}$ is $r$-uniformly dense on compacta in
	\begin{align*}
	\left\{\Phi: \R_0^+ \times U \to \R^{2d} \mid \Phi \text{ is phase flow of \cref{eqn:Hsys}}\right\},
	\end{align*}
	where $H$ is given by
	\begin{align*}
	H(p,q)= K(p) + V(q)
	\end{align*}
	for some $K,V \in C^{r+1}(U),$ if the activation function $\sigma$ is $r$-finite.
\end{theorem}
\begin{proof}
    First, we want to show that for all $\epsilon > 0$, there are $m \in \N$ and $V_1,\ldots,V_{4m}\in C^{r+1}$ such that
    \begin{align}\label{eq:difhenonapr}
        \norm{\Phi(h,\cdot) - \begin{bmatrix}
            h/m \nabla V_{4m} & -I\\ I & 0
        \end{bmatrix}\circ \cdots \circ
        \begin{bmatrix}
            h/m \nabla V_{1} & -I\\ I & 0
        \end{bmatrix}}_{C^r([0,T]\times W; \R^{2d})}\leq\epsilon.
    \end{align}
    To do this, we define $V_{4k+1}\equiv 0,~V_{4k+2}(x)=V(-x),~V_{4k+3}(x)=K(-x),~V_{4k+4}\equiv 0$ for $k \in \N_0$ and observe
    \begin{align*}
        g_{\frac{h}{m}}\begin{pmatrix}
            p \\ q
        \end{pmatrix}
        :=&\begin{bmatrix}
            h/m \nabla V_{4k+4} & -I\\ I & 0
        \end{bmatrix}
        \begin{bmatrix}
            h/m \nabla V_{4k+3} & -I\\ I & 0
        \end{bmatrix}
        \begin{bmatrix}
            h/m \nabla V_{4k+2} & -I\\ I & 0
        \end{bmatrix}
        \begin{bmatrix}
            h/m \nabla V_{4k+1} & -I\\ I & 0
        \end{bmatrix}
        \begin{pmatrix}
            p \\ q
        \end{pmatrix}\\
        =&
        \begin{pmatrix}
            p - h/m \nabla V(q)\\
            q + h/m \nabla K(p - h/m \nabla V(q))
        \end{pmatrix}.
    \end{align*}
    This implies
    \begin{align*}
        \norm{g_m-\Phi_{1,m}}_{C^r([0,T]\times \tilde{W}_\Phi, \R^{2d})} \leq \frac{C}{m^2}
    \end{align*}
    for some $C< \infty$, according to the proof of \cite[Theorem 4]{JanB25}. Now \Cref{the:comp} yields \eqref{eq:difhenonapr} if we choose $m>\tilde{C}/\epsilon$.
    This means we have $\normt{\Phi -f_{4m}\circ \cdots \circ f_1}\leq\epsilon$ for 
    \begin{align*}
        f_i\left(h, \begin{pmatrix}
            p \\ q
        \end{pmatrix}\right):=
        \begin{bmatrix}
            h/m \nabla V_i & -I \\ I & 0
        \end{bmatrix}
        \begin{pmatrix}
            p\\ q
        \end{pmatrix},
    \end{align*}
    $i=1,\ldots,4m.$\\
    Now we construct $V_0, \eta_0$ such that $f_{4m}\circ \cdots \circ f_1$ is the same as applying a certain $\mathcal{H}_{V_0,\eta_0}$ $4m$-times by adapting the proof of \cite[Theorem 1]{Tur03}. Without loss of generality, let $W \ni 0$. Furthermore, let
    \begin{align*}
        L:= \max_{i=1,...,4m}\max_{h \in [0,T]} \max_{x \in W} \norm{(f_i(h,\cdot)\circ \cdots \circ f_1(h,\cdot))(x)}_\infty,
    \end{align*}
    and we choose $\mu:=\eta_0^{(q)}=\eta_0^{(p)}/2$ and $\norm{\mu}_\infty > 2L$.
    We define $p_0^*=0,~p_1^*=2\mu,~p_2^*=\mu,~p_3^*=-\mu$ and observe that $\norm{p^*_{i}-p^*_{j}}\geq \mu > 2L$ for $i,j=1,\ldots,4,~i\neq j$. Therefore, 
    \begin{align*}
        V_0(p)=\sum_{i=1}^{4} \chi_{\overline{B(p_{i-1}^*, L)}}(p)\left[\frac{1}{m}V_i(p - p_{i-1}^*)\right],
    \end{align*}
    with $\chi_A$ being the characteristic function of $A$, always only evaluates one $V_i$ for $i=1,\ldots,4$. It is easy to verify that 
    \begin{align*}
        \mathbf{H}_{V_0,\eta_0}(h,\cdot)=f_4(h,\cdot) \circ f_3(h,\cdot) \circ f_2(h,\cdot) \circ f_1(h,\cdot).
    \end{align*}
    Because $V_{4k+i}=V_i$ for $k=1,...,m$ and $i=1,...,4$, this yields
    \begin{align*}
        \underbrace{\mathbf{H}_{V_0, \eta_0}(h,\cdot)\circ \hdots \circ \mathbf{H}_{V_0, \eta_0}(h,\cdot)}_{m\text{-times}}=f_{4m}(h,\cdot) \circ \cdots \circ f_1(h,\cdot).
    \end{align*}
    Since $V_0$ can be approximated with arbitrary accuracy by a neural network  with one hidden layer $V_\Theta$ \cite{HorSW90}, we get\\$\normtd{\mathcal{H}_{V_0,\eta_0}-\mathcal{H}_{V_\Theta,\eta_\Theta}}<\epsilon$ analogous to \cite[Lemma 3]{JinZZetal20}. Now the theorem can be obtained by following the proof of \cite[Theorem 5]{JinZZetal20}
\end{proof}

\begin{theorem}[Approximation theorem for non-autonomous T-HénonNets]
    \label{the:nathenonnet}
	For any $r \in \mathbb{N}$ and open $U \subseteq \R^{2d},$ the set of NAT-HénonNets $\Psi_\text{NATH}$ is $r$-uniformly dense on compacta in
	\begin{align*}
	\left\{\Phi: \R_0^+ \times \R_0^+\times U \to \R^{2d} \mid \Phi \text{ is phase flow of \cref{eqn:Hsys}}\right\},
	\end{align*}
	where $H$ is given by
	\begin{align*}
	H(p,q,t)= K(p,t) + V(q,t)
	\end{align*}
	for some $K,V \in C^{r+1}(U\times \R_0^+),$ if the activation function $\sigma$ is $r$-finite.
\end{theorem}
\begin{proof}
    Analogous to the proof of \Cref{the:henonnet}. Since the system is non-autonomous, we have to use \Cref{the:compcor} instead of \Cref{the:comp}.
\end{proof}

\subsection{Limitations}
\label{subsec:lim}
In this subsection, we discuss the limitations of T-HénonNets. As promised in \Cref{subsec:thenonnets}, we start by showing that the naive T-HénonNets based on \eqref{eqn:naiv_henon} are limited to systems with Hamiltonians of the form $H(p, q)= V(-p)+ V(q)$. After that, we show that T-HénonNets cannot deal with non-separable Hamiltonian systems.\\
To get both of these results, we observe that in general for the flow $\Phi$ it holds
\begin{align*}
	\partial_t \Phi(0,x)= \dot{x}= J^{-1}\nabla H(x).
\end{align*}
So we can use $\partial_h \psi(0,x)$ to determine which Hamiltonian systems a T-HénonNet can approximate. Therefore, it is useful to observe
\begin{align*}
	\partial_h \psi(0,x)&=\frac{\partial \mathbf{H}_{V_m,\eta_m}}{\partial x}(0,(\mathbf{H}_{V_{m-1},\eta_{m-1}}\circ\cdots \circ \mathbf{H}_{V_1,\eta_1})(0,x))\frac{d}{dh}(\mathbf{H}_{V_{m-1},\eta_{m-1}}\circ\cdots\circ \mathbf{H}_{V_1,\eta_1})(0,x)\notag \\
    & \qquad+ \frac{\partial \mathbf{H}_{V_m,\eta_m}}{\partial h}(0,(\mathbf{H}_{V_{m-1},\eta_{m-1}}\circ\cdots \circ \mathbf{H}_{V_1,\eta_1})(0,x))\\
	&=\begin{pmatrix}
	I & 0\\
	0 & I
	\end{pmatrix}\frac{d}{dh}(\mathbf{H}_{V_{m-1},\eta_{m-1}}\circ\cdots\circ \mathbf{H}_{V_1,\eta_1})(0,x)+\frac{\partial \mathbf{H}_{V_m,\eta_m}}{\partial h}(0,x), \notag
\end{align*}
which holds because $\mathbf{H}_{V_m,\eta_m}(0,x)=x$. By repeating this argument, we obtain
\begin{align}
    \label{eqn:dh}
	\partial_h \psi(0,x)=\sum_{i=1}^{m}\frac{\partial \mathbf{H}_{V_i, \eta_i}}{\partial h}(0,x).
\end{align}
With this preparation, we are ready to take a closer look at why the limitations of the naive T-HénonNets.
\subsubsection{Naive T-HénonNets}
We recall \eqref{eqn:naiv_henon}, i.e.,
\begin{align*}
    \mathcal{H}_{V,\eta}\left(h, \pq\right)=\begin{bmatrix}
        h\nabla V & -I\\ I & 0
    \end{bmatrix}\pq + \begin{pmatrix}
        0 \\ \eta
    \end{pmatrix}
\end{align*}
and observe
\begin{align*}
    \frac{\partial \mathbf{H}_{V, \eta}}{\partial h}\left(0,\pq\right)=\begin{pmatrix}
        - \nabla V (-q) + \nabla V(q - \eta)\\
        - \nabla V(p) + \nabla V(-p-\eta)
    \end{pmatrix}.
\end{align*}
Using \eqref{eqn:dh}, we have
\begin{align*}
    \partial_h \psi\left(0,\pq\right)=\sum_{i=1}^{m}\begin{pmatrix}
        - \nabla V_i (-q) + \nabla V_i(q - \eta_i)\\
        - \nabla V_i(p) + \nabla V_i(-p-\eta_i)
    \end{pmatrix}= \begin{pmatrix}
        - \nabla V(q)\\ - \nabla V(-p)
    \end{pmatrix}= J^{-1}\begin{pmatrix}
        - \nabla V(-p)\\ \nabla V(q)
    \end{pmatrix},
\end{align*}
with $V(y):=\sum_{i=1}^m -V_i(-y) - V_i(y-\eta_i)$. So the associated Hamiltonian $\tilde{H}$ is given by $\tilde{H}(p,q)=V(-p) +  V(q)$.

\subsubsection{Non-separable Hamiltonian systems}
In this subsection, we show that T-HénonNets cannot deal with non-seprable Hamiltonian systems. 
To show that these limitations were not introduced by \Cref{def:thenonnet}, we generalize their definition. We allow $\eta = \eta(t)$ and $V=V(h,p)$ to be time-dependent as well as different potentials $V$ and shifts $\eta$ in each layer, i.e.,
\begin{align*}
    \mathbf{H}_i&=\mathcal{H}_{V_{4i},\eta_{4i}}\circ\mathcal{H}_{V_{4(i-1)+3},\eta_{4(i-1)+3}}\circ\mathcal{H}_{V_{4(i-1)+2},\eta_{4(i-1)+2}}\circ\mathcal{H}_{V_{4(i-1)+1},\eta_{4(i-1)+1}}\\
    \psi&=\mathbf{H}_m \circ \cdots \circ \mathbf{H}_1.
\end{align*}
Since we want to strongly enforce $\psi(0,x)=x$, we implement $\mathbf{H}_{i}(0,x)=x$ by assuming $\nabla V_i(0,x)\equiv 0$ for $i=1,\ldots,4m$ and $\eta_{4k+1}(0)=\eta_{4k+3}(0),~ \eta_{4k+2}(0)=\eta_{4k+4}(0)$ for $k=0,\ldots,m-1.$ Under these assumptions, we observe
\begin{align*}
    \frac{\partial \mathbf{H}_i}{\partial h}(0,x)&=\begin{pmatrix}
        \partial_h \nabla V_{4i}(0,q-\eta_{4(i-1)+2}^{(q)}(0))-\partial_h \nabla V_{4(i-1)+2}(0,-q + \eta_{4(i-1)+1}^{(p)}(0))\\
        \partial_h \nabla V_{4(i-1)+3}(0,-p-\eta_{4(i-1)+1}^{(q)}(0)+\eta_{4(i-1)+2}^{(p)}(0)) - \partial_h \nabla V_{4(i-1)+1}(0,p)
    \end{pmatrix}\\
    & \qquad + \begin{pmatrix}
        \dot{\eta}_{4(i-1)+1}^{(q)}(0)-\dot{\eta}^{(q)}_{4(i-1)+3}(0) - \dot{\eta}_{4(i-1)+2}^{(p)}(0) + \dot{\eta}_{4i}^{(p)}(0)\\
        - \dot{\eta}^{(q)}_{4(i-1)+2}(0)+\dot{\eta}^{(q)}_{4i}(0) - \dot{\eta}^{(p)}_{4(i-1)+1}(0) + \dot{\eta}_{4(i-1)+3}^{(p)}(0)
    \end{pmatrix}.
\end{align*}
Using \cref{eqn:dh} again, we get
\begin{align}
	\label{eqn:inf_sep}
    \partial_h \psi(0,x)=\sum_{i=1}^m\frac{\partial u_i}{\partial t}(0,x)
    =\begin{pmatrix}
        -\nabla V(q)\\ \nabla K(p)
    \end{pmatrix},
\end{align}
with
\begin{align*}
    V(q)&:=-\sum_{k=1}^m \partial_h V_{4i}(0,q-\eta_{4(i-1)+2}^{(q)}(0))+\partial_h V_{4(i-1)+2}(0,-q+ \eta_{4(i-1)+1}^{(p)}(0))\\
    & \qquad\qquad+(\dot{\eta}_{4(i-1)+1}^{(q)}(0)-\dot{\eta}^{(q)}_{4(i-1)+3}(0) - \dot{\eta}_{4(i-1)+2}^{(p)}(0) + \dot{\eta}_{4i}^{(p)}(0))^T q,\\
    K(p)&:=\sum_{k=1}^m -\partial_h  V_{4(i-1)+3}(0,-p-\eta_{4(i-1)+1}^{(q)}(0)+\eta^{(p)}_{4(i-1)+2}(0)) - \partial_h  V_{4(i-1)+1}(0,p)\\
    &\qquad\qquad+ (- \dot{\eta}^{(q)}_{4(i-1)+2}(0)+\dot{\eta}^{(q)}_{4i}(0) - \dot{\eta}^{(p)}_{4(i-1)+1}(0) + \dot{\eta}_{4(i-1)+3}^{(p)}(0))^T p.
\end{align*}
So even in this very general formulation, we have $\tilde{H}=K(p)+V(q)$ and therefore, observe that T-HénonNets are not able to represent the flow of non-separable Hamiltonian systems.
\section{Numerical examples}
\label{sec:exp}
In this section we validate our theoretical results numerically. We choose the same setup as in \cite{JanB25} to make T-HénonNets comparable to TSympNets. In \Cref{subsec:pend}, we verify that T-HénonNets can solve the pendulum problem, since this is a popular example in many machine learning papers \cite{JanB25,JinZZetal20, BurTM21}. In \Cref{subsec:lin}, we show that the theoretical limitation of the presented method to separable Hamiltonian systems also holds in practice. We consider a very easy linear but non-separable Hamiltonian system and observe that T-HénonNets fail at learning the system. In \Cref{subsec:ho}, we show the necessity of our extensions for non-autonomous Hamiltonian systems. We observe that the autonomous T-HénonNet fails, while the non-autonomous method is able to make accurate predictions.\\
The training in all following subsections is performed by minimizing the mean-squared error (MSE)
\begin{align*}
    L = \frac{1}{N}\sum_{i=1}^N\norm{\psi(x_i)-y_i}_2^2
\end{align*}
for $x_i,y_i$ in the training set $\mathcal{T}=\{(x_i,y_i)\}_{i=1}^N$ with the Adam optimizer \cite{KinB17}. We use $\tanh$ as our activation function and the architectures found in \Cref{tab:arch}. To perform the training, we adapted the "Learner" framework.\footnote{\texttt{\url{https://github.com/jpzxshi/learner}}}
\begin{table}
    \centering
    \begin{tabular}{|c|c|c|c|c|}
     \hline {Experiment} & NN Type & Layers & Width & Parameters \\  \hline\hline {Pendulum} & T-HénonNet & 5 & 30 & 460 \\ \hline {Linear example} & T-HénonNet & 5 & 30 & 460 \\   \hline \multirow{2}{*}{Forced harmonic oscillator} & T-HénonNet & 6  & 20 & 372 \\  \cline{2-5}&  NAT-HénonNet & 6 & 20 & 492 \\\hline \end{tabular}
     \caption{Architecture of the HénonNets used in the different experiments}
     \label{tab:arch}
\end{table}

\subsection{Pendulum}
\label{subsec:pend}
\begin{figure}
    \centering
    \begin{subfigure}{0.5\textwidth}
        \includegraphics[width=0.9\linewidth]{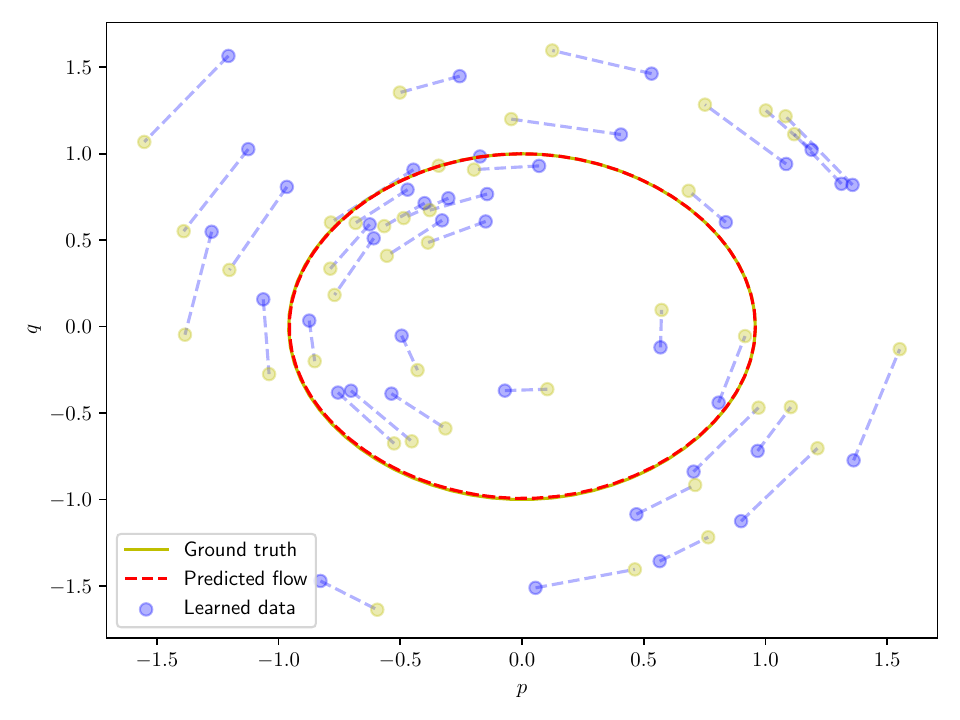}
        \caption{Phase flow}
    \end{subfigure}%
    \begin{subfigure}{0.5\textwidth}
        \includegraphics[width=0.9\linewidth]{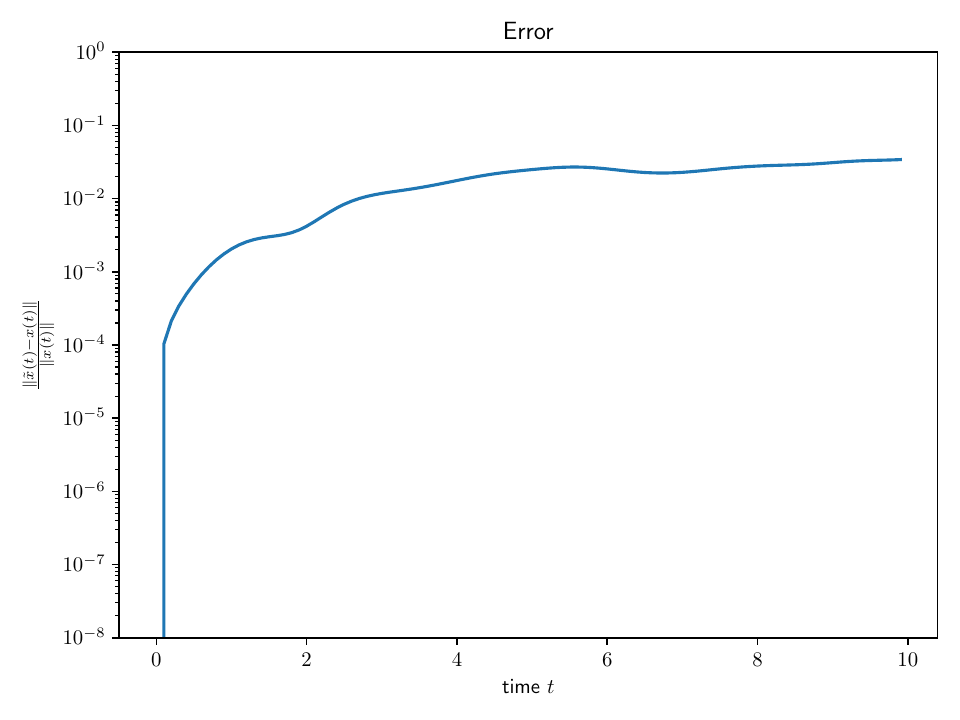}
        \caption{Error}
    \end{subfigure}
    \caption{Pendulum example $H(p,q)=\frac{1}{2}p^2- \cos(q)$. Left: Phase flow, blue and yellow points with dashed lines represent the training data. The solid yellow line is the test trajectory. Right: Relative error of the prediction compared to the test trajectory.}
    \label{fig:pend}
\end{figure}

We first consider the pendulum system with
\begin{align*}
    H(p,q)= \frac{1}{2}p^2 - \cos(q)
\end{align*}
as it is a widely used example for symplectic machine learning methods. Also, we use the same kind of training and test datasets as in \cite{JinZZetal20} and \cite{JanB25}, i.e., the training dataset consists of $N=40$ tuples $\mathcal{T}=\{([x_i,h_i],y_i)\}_{i=1}^N$. The phase space coordinates $\{x_i\}_{i=1}^N$ are sampled from a uniform distribution over $[-\sqrt{2}, \sqrt{2}]\times [-\frac{1}{2}\pi, \frac{1}{2}\pi]$. The time steps $\{h_i\}_{i=1}^N$ are sampled from a uniform distribution over $[0.2,0.5]$. The labels $y_i$ are predictions of the flow $y_i=\Phi(h_i,x_i)$. Since we do not have an analytical representation of the flow $\Phi$, we use the 6th-order Störmer-Verlet Scheme \cite{HaiLW03} and make 10 steps of size $h_i/10$ to obtain $y_i$. The test dataset is a single trajectory with $k=100$ fixed time steps $h=0.1$ starting at $x_0=(1,0)^T$. All neural networks are trained for 50000 epochs with a learning rate of $10^{-3}$. We test T-HénonNets with the hyper parameters found in \Cref{tab:arch}. Their prediction is generated by making $k$ forward evaluations of the respective neural network.
As we can see in \Cref{fig:pend}, T-HénonNets are able to capture the dynamics of the pendulum.

\subsection{Linear example}
Now we consider a simple non-separable linear example with the Hamiltonian given by
\begin{align*}
    H(p, q)=\frac{1}{2}p^2 + 0.4 pq + \frac{1}{2}q^2.
\end{align*}
The point of this experiment is to show that the limitations found in \Cref{subsec:lim} are not only of theoretical nature, but also have practical implications. Recall that T-HénonNets should not be able to deal well with non-separable Hamiltonian systems, because they always infer separable Hamiltonians according to \eqref{eqn:inf_sep}.\\
Training and test datasets are generated the same way as in the pendulum example in \Cref{subsec:pend}. The used architectures can be found in \Cref{tab:arch} again.
\label{subsec:lin}
\begin{figure}
    \centering
    \begin{subfigure}{0.5\textwidth}
        \includegraphics[width=0.9\linewidth]{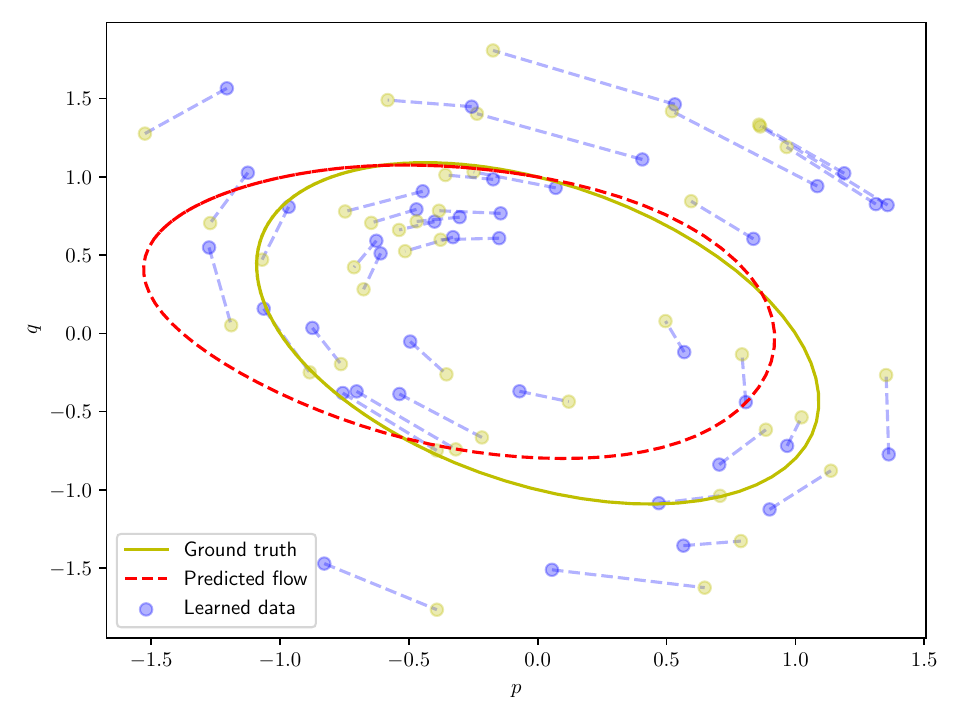}
        \caption{Phase flow}
    \end{subfigure}%
    \begin{subfigure}{0.5\textwidth}
        \includegraphics[width=0.9\linewidth]{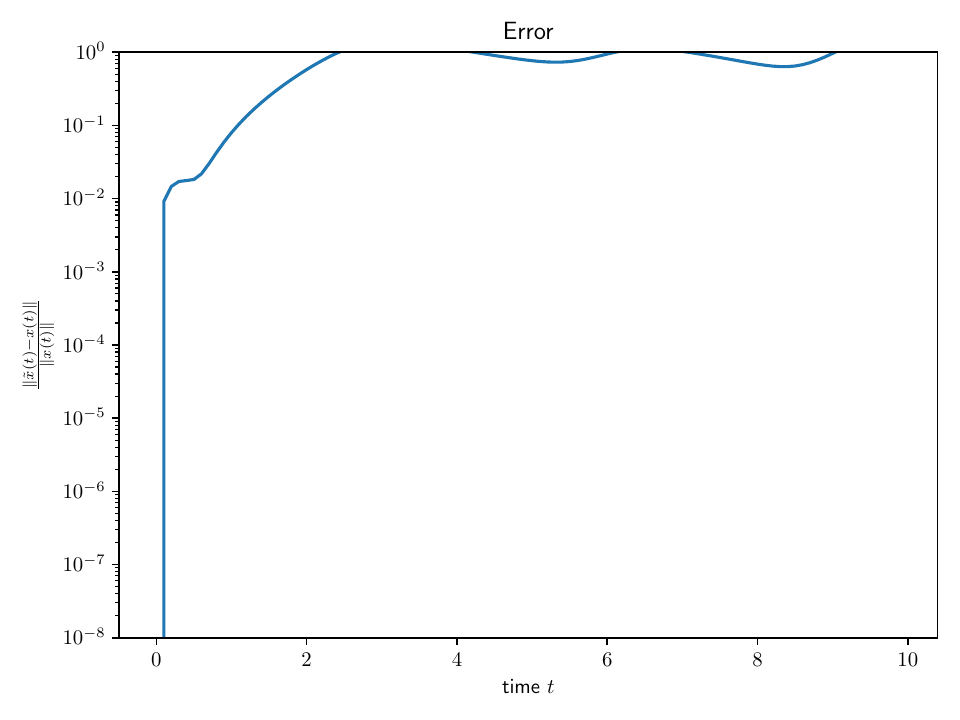}
        \caption{Error}
    \end{subfigure}
    \caption{Linear example $H(p,q)=\frac{1}{2}p^2 + 0.4 pq + \frac{1}{2} q^2$. Left: Phase flow, blue and yellow points with dashed lines represent the training data. The solid yellow line is the test trajectory. Right: Relative error of the prediction compared to the test trajectory}
    \label{fig:lin}
\end{figure}
In \Cref{fig:lin}, we observe that our theoretical results from \Cref{subsec:lim} hold true in practice. T-HénonNets fail to capture the dynamics of the non-separable linear example correctly.

\subsection{Forced harmonic oscillator}
\label{subsec:ho}
\begin{figure}
    \centering
    \begin{subfigure}{.5\textwidth}
      \centering
      \includegraphics[width=.9\linewidth]{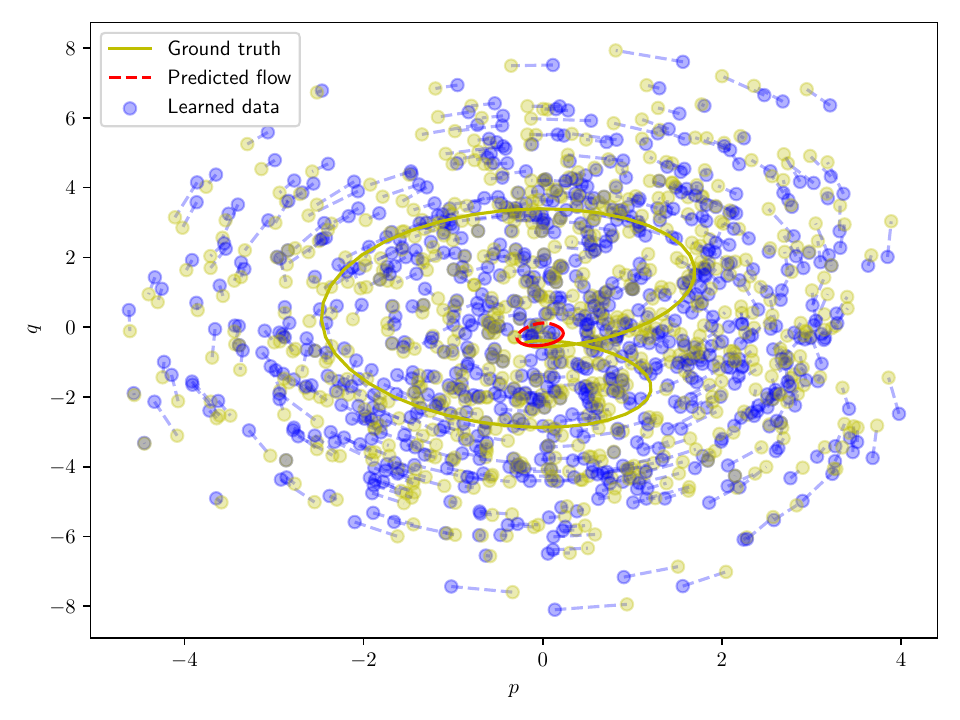}
      \caption{Phase flow T-HénonNet}
      \label{fig:ho_TH_flow}
    \end{subfigure}%
    \begin{subfigure}{.5\textwidth}
      \centering
      \includegraphics[width=.9\linewidth]{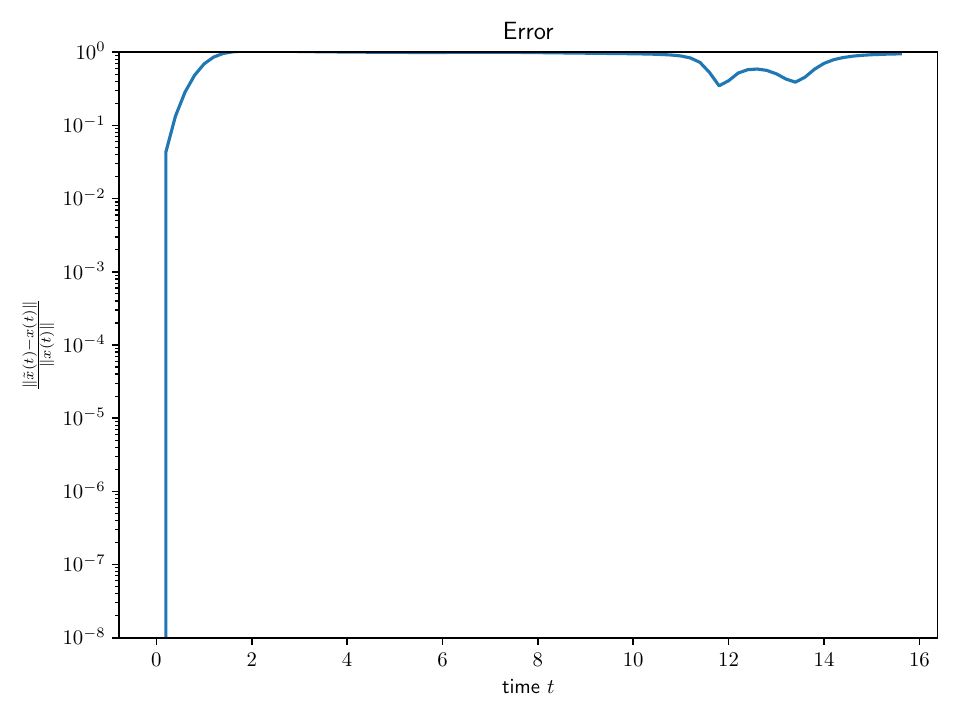}
      \caption{Error T-HénonNet}
      \label{fig:ho_TH_err}
    \end{subfigure}
    \begin{subfigure}{.5\textwidth}
        \centering
        \includegraphics[width=0.9\linewidth]{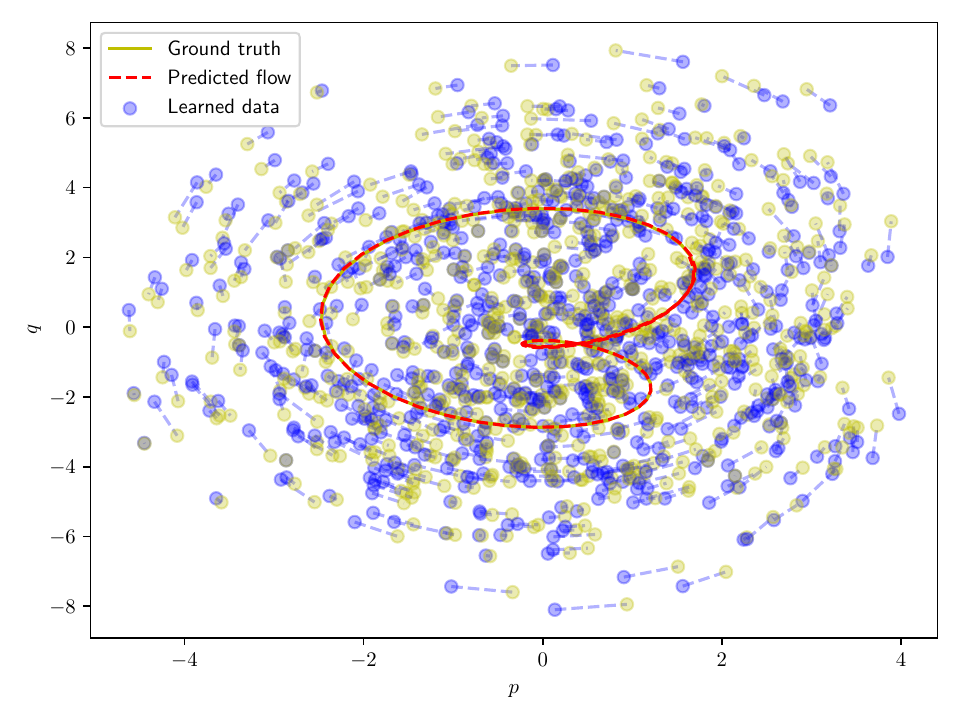}
        \caption{Phase flow NAT-HénonNet}
        \label{fig:ho_NATH_flow}
    \end{subfigure}%
    \begin{subfigure}{.5\textwidth}
        \centering
        \includegraphics[width=0.9\linewidth]{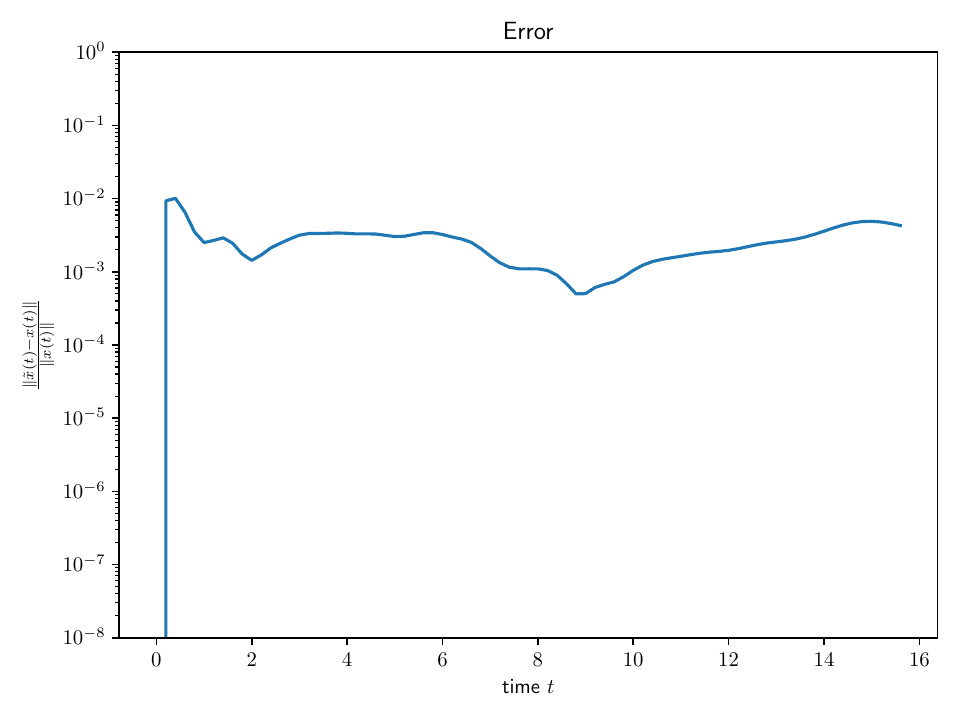}
        \caption{Error NAT-HénonNet}
        \label{fig:ho_NATH_err}
    \end{subfigure}
    \caption{Forced harmonic oscillator $H(p,q,t)=\frac{1}{2}p^2+\frac{1}{2}\omega_0^2 q^2 - F_0 \sin(\omega t)q$. Left: Phase flow, blue and yellow points with dashed lines represent the training data. The solid yellow line is the test trajectory. Left: Relative error of the prediction compared to the test trajectory}
    \label{fig:ho}
\end{figure}

As our last example, we consider the harmonic oscillator with an external force. The non-autonomous Hamiltonian is given by
\begin{align*}
    H(p, q, t)=\frac{1}{2}p^2+\frac{1}{2}\omega_0^2 q^2 - F_0 \sin(\omega t)q.
\end{align*}
Fortunately, the corresponding Hamiltonian system can be solved analytically. The analytical solution is given by
\begin{align*}
    p(t)&=\left(p_0 -\frac{\omega F_0}{\omega_0^2 - \omega^2}\right)\cos(\omega_0 t) -q_0 \omega_0 \sin(\omega_0 t)+ \frac{\omega F_0}{\omega_0^2 - \omega^2}\cos(\omega t),\\
    q(t)&=q_0 \cos(\omega_0 t) + \left(\frac{p_0}{\omega_0} -\frac{\omega F_0}{\omega_0(\omega_0^2 - \omega^2)}\right)\sin(\omega_0 t) + \frac{F_0}{\omega_0^2 - \omega^2}\sin(\omega t),
\end{align*}
with initial conditions $p(0)=p_0,~q(0)=q_0$. Hence, we do not need to use a numerical solver to generate our training and test data. The training data set is given by $N=800$ tuples $\mathcal{T}=\{[x_i,t_i,h_i],[y_i,\tilde{t}_i\}_{i=1}^N$, where $\{x_i\}_{i=1}^N$ is sampled from a uniform distribution over $[-3.5,2]\times[-4,4]$. The times $\{t_i\}_{i=1}^N$ and time steps $\{h_i\}_{i=1}^N$ are also sampled from uniform distributions over $[0,16]$ and $[0,0.3]$ respectively. The labels $\{y_i\}_{i=1}^N$ are given by 
$y_i=(p(t_i+h_i)^T, q(t_i+h_i)^T)^T$. The test data is given by the trajectory of the analytical solution with initial value $(p_0,q_0)=(-0.2,-0.5)$ evaluated at equidistant times $t_i=ih$ with $h=0.2$ for $i=1,\ldots,k,~k=80$. For this example, we test T-HénonNets and NAT-HénonNets with architectures from \Cref{tab:arch} to show the necessity of the NAT-HénonNets. Both neural networks are trained for 40000 epochs with a learning rate of $10^{-3}$.\\
\Cref{fig:ho} shows that both T-HénonNets are not able to capture the dynamics of the harmonic oscillator with an external force. This was expected, because the test trajectory intersects with itself. So the T-HénonNet would need to give different predictions at the same phase space coordinate $(p,q)$ depending on the time $t$. Since time is not an input of T-HénonNets, this is impossible.\\
\Cref{fig:ho_NATH_flow,fig:ho_NATH_err} show that NAT-HénonNets are able to capture the dynamics well. The mentioned intersection is not a problem for NAT-HénonNets, since they live on the $(p,q,t)$ phase space and the intersection only happens in the $(p,q)$ space.\\
\section*{Code and Data Availability Statement}
\label{sec:code}
The relevant code and data for this work has been archived within the Zenodo repository \cite{JanB25software}.
\section{Conclusions and outlook}
\label{sec:conc}
In this work, we developed a new time-adaptive symplectic neural network architecture, which we coined T-HénonNets and established a universal approximation theorem for them limited to separable Hamiltonian systems. This limitation is not an artifact of our proofs, but a direct result of the used architecture shown in \Cref{subsec:lim} theoretically and \Cref{subsec:lin} experimentally. Additionally, we proposed a novel intrinsically symplectic, time-adaptive neural network architecture to handle non-autonomous Hamiltonian systems in NAT-HénonNets. We were able to show that they are universal approximators for the flows of separable non-autonomous Hamiltonian systems. Lastly, we verified our theoretical findings on three numerical experiments.\\
The obvious open research question is how to adapt T-HénonNets such that they will be able to handle non-separable Hamiltonian systems as well.
%%%%%%%%%%%%%%%%%%%%%%%%%%%%%%%%%%%%%%%%%%%%%%%%%%%%%%%%%%%%%%%%%%%%%%%%%%%%%%%%
% *** REFERENCES ***                                                           %
%%%%%%%%%%%%%%%%%%%%%%%%%%%%%%%%%%%%%%%%%%%%%%%%%%%%%%%%%%%%%%%%%%%%%%%%%%%%%%%%

% Use this code if you wish to generate your bibliography with BibTeX;
% please replace first the string "demo" below with the name(s) of
% the BibTeX data base(s) you want to use.
% The resulting bibliography-output (the contents of the .bbl file)
% must be pasted into this file before submission.
% 
\bibliographystyle{pamm}
\bibliography{ref}

\end{document}